\pdfoutput=1

\documentclass[11pt]{article}

\usepackage{EMNLP2023}

\usepackage{times}
\usepackage{latexsym}
\usepackage{graphicx}
\usepackage{amsmath}
\usepackage{multirow}
\usepackage{subfig}

\usepackage[T1]{fontenc}

\usepackage[utf8]{inputenc}

\usepackage{microtype}

\usepackage{inconsolata}
\usepackage{booktabs,makecell}

\title{Beyond Fine-tuning: Unleashing the Potential of Continuous Pretraining for Clinical LLMs.\Thanks{This work has been accepted at EMNLP Findings 2024.}}

\author{Cl{\'e}ment Christophe, Tathagata Raha, Svetlana Maslenkova, Muhammad Umar Salman, \\
\textbf{Praveen K Kanithi, Marco AF Pimentel, Shadab Khan} \\
M42 Health, Abu Dhabi, UAE\\
\texttt{cchristophe@m42.ae}}

\begin{document}
\maketitle
\begin{abstract}
Large Language Models (LLMs) have demonstrated significant potential in transforming clinical applications. In this study, we investigate the efficacy of four techniques in adapting LLMs for clinical use-cases: continuous pretraining, instruct fine-tuning, NEFTune \cite{jain2023neftune}, and prompt engineering. We employ these methods on Mistral 7B and Mixtral 8x7B models, leveraging a large-scale clinical pretraining dataset of 50 billion tokens and an instruct fine-tuning dataset of 500 million tokens. Our evaluation across various clinical tasks reveals the impact of each technique. While continuous pretraining beyond 250 billion tokens yields marginal improvements on its own, it establishes a strong foundation for instruct fine-tuning. Notably, NEFTune, designed primarily to enhance generation quality, surprisingly demonstrates additional gains on our benchmark. Complex prompt engineering methods further enhance performance. These findings show the importance of tailoring fine-tuning strategies and exploring innovative techniques to optimize LLM performance in the clinical domain.
\end{abstract}

\section{Introduction}
 
The advent of large language models (LLMs) has spurred a wave of innovation across various domains, with healthcare being a particularly promising area for their application. LLMs have the potential to transform clinical workflows, aid in diagnosis, and enhance patient care. However, effectively adapting these models to the nuances and complexities of the clinical domain remains a significant challenge.

Current approaches in the literature predominantly focus on either developing specialized clinical LLMs from scratch or fine-tuning existing models on large-scale clinical datasets. While these methods have shown promise, they often overlook the potential benefits of continuous pretraining on domain-specific data as a means to further enhance model performance. This is due in part to the complexities and potential instabilities associated with continued training of large models.

In this study, we take a comprehensive approach to optimizing clinical LLMs by systematically investigating the impact of continuous pretraining on in-domain data, in conjunction with instruct fine-tuning and advanced prompting strategies. We focus on the Mistral-7B \cite{jiang2023mistral} and Mixtral-8x7B \cite{jiang2024mixtral} models, demonstrating that continuous pretraining, while yielding modest gains compared to fine-tuning and prompting, plays a crucial role in establishing a solid foundation for further specialization. By carefully balancing in-domain clinical data with general language data, we successfully mitigate instability issues and unlock the full potential of continuous pretraining for clinical LLMs.

Our work highlights the importance of understanding of the relationship between pretraining, fine-tuning, and prompting in adapting LLMs for clinical applications. By demonstrating the effectiveness of continuous pretraining on domain-specific data, we open doors for future research to further explore this underutilized technique to develop more accurate, reliable, and ultimately impactful clinical LLMs.

\section{Related Works}

The landscape of Large Language Models (LLMs) for healthcare is evolving rapidly, with most approaches involving either domain-specific pretraining or instruction fine-tuning of general-purpose models. OpenAI's GPT-3.5 and GPT-4 \cite{openai2023gpt4}, alongside Google's Med-PaLM \cite{singhal2023large} and Med-PaLM 2 \cite{singhal2023towards} have demonstrated impressive performance on medical benchmarks, despite limited transparency regarding their training details. Other models, such as GatorTron \cite{Yang2022-kt}, and PMC-LLaMA \cite{Wu2023-sx}, have shown the potential of pretraining on extensive biomedical corpora to add domain-specific knowledge for clinical applications.

Instruction fine-tuning and dialogue datasets have also been instrumental in enhancing the zero-shot and few-shot generalization capabilities of LLMs. ChatDoctor \cite{Li2023-ro} and MedAlpaca \cite{Han2023medalpaca}, for instance, utilize medical conversations and other NLP tasks to improve LLaMA's performance on clinical queries. Recent models like Clinical Camel \cite{toma2023clinical}, MediTron \cite{chen2023meditron}, HuatuoGPT-2 \cite{chen2023huatuogpt} and Med42 \cite{christophe2024med42}, based on LLaMA-2 \cite{touvron2023llama2}, further demonstrate the efficacy of this approach.

Building on the observation that models can learn from prompting alone \cite{brown2020language}, recent research has explored techniques to enhance clinical capabilities without additional training. These methods often extend the well-known Chain-of-Thought prompting technique, originally introduced by \cite{wei2022chain}, to better suit clinical use-cases. Notably, Microsoft's MedPrompt \cite{nori2023can} demonstrates significant improvements in GPT-4's performance on clinical QA tasks, while \cite{garikipati2024openmedlm} apply similar strategies to the Yi family of models \cite{young2024yi}. Google has also showcased the potential of complex prompting to boost the clinical capabilities of their Gemini model \cite{saab2024capabilities}. However, while such complex prompting techniques can improve performance on standard benchmarks, their practicality and scalability in real-world clinical applications remain to be seen.

Recent studies like LIMA \cite{zhou2024lima}, FineWeb \cite{fineweb2024} and Phi \cite{li2023textbooks} have highlighted the pivotal role of data quality in LLM training, emphasizing that it can often be more influential than architectural choices in determining model performance. High-quality data has been shown to significantly impact the model's ability to learn meaningful representations and generalize to new tasks. This shows the importance of our approach to dataset curation, ensuring that our models are trained on a robust and representative collection of clinical data.

\section{Experiments}

In this section, we present the four steps of our experimental framework: (1) continuous pretraining, (2) instruct fine-tuning, (3) NEFTune, and (4) complex prompt engineering.

\subsection{Continuous Pretraining}

Continuous pretraining involves extending the pretraining phase of a large language model (LLM) by exposing it to additional text data. This can be particularly beneficial in domain-specific applications, like healthcare, where models can be further trained on vast amounts of clinical literature. The goal is to refine the model's understanding of domain-specific terminology, relationships, and nuances, potentially leading to improved performance on relevant tasks. In our experiments, we investigate the impact of continuous pretraining on both Mistral 7B and Mixtral 8x7B models, utilizing a 50-billion-token clinical dataset.

Continuous pretraining of large language models, however, is not without its challenges. Typically, only the weights of the LLM are openly accessible, while the optimizer state remains unavailable. This lack of access can disrupt the training process, leading to instabilities and hindering the model's ability to effectively learn from the new data. Additionally, the potential distribution shift between the original pretraining data and the new clinical data can result in catastrophic forgetting, where the model loses proficiency on previously learned knowledge and tasks \cite{li2024examining}.

\begin{figure*}
\centering
\subfloat{\includegraphics[width=0.5\textwidth]{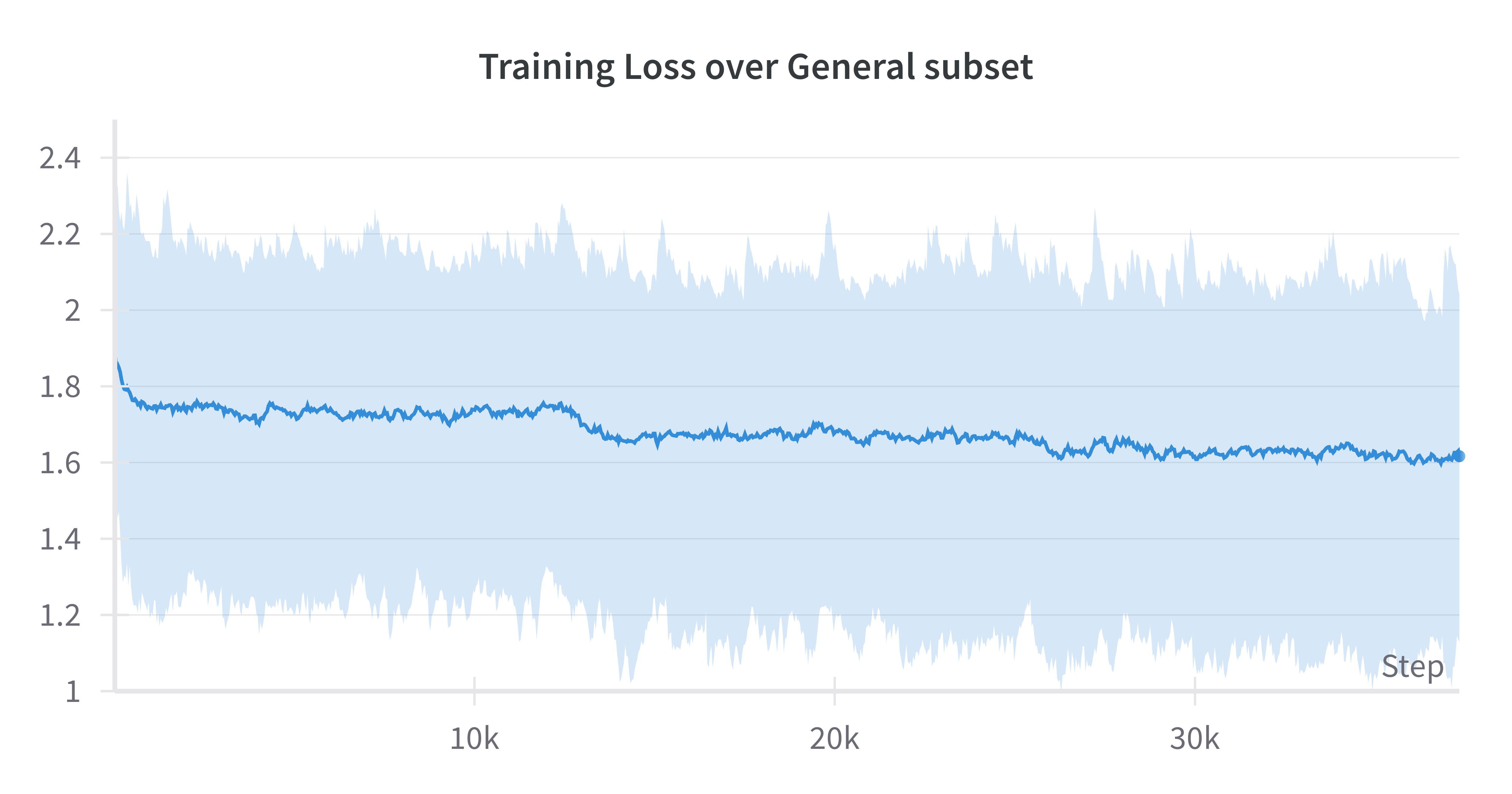}}
\hfill
\subfloat{\includegraphics[width=0.5\textwidth]{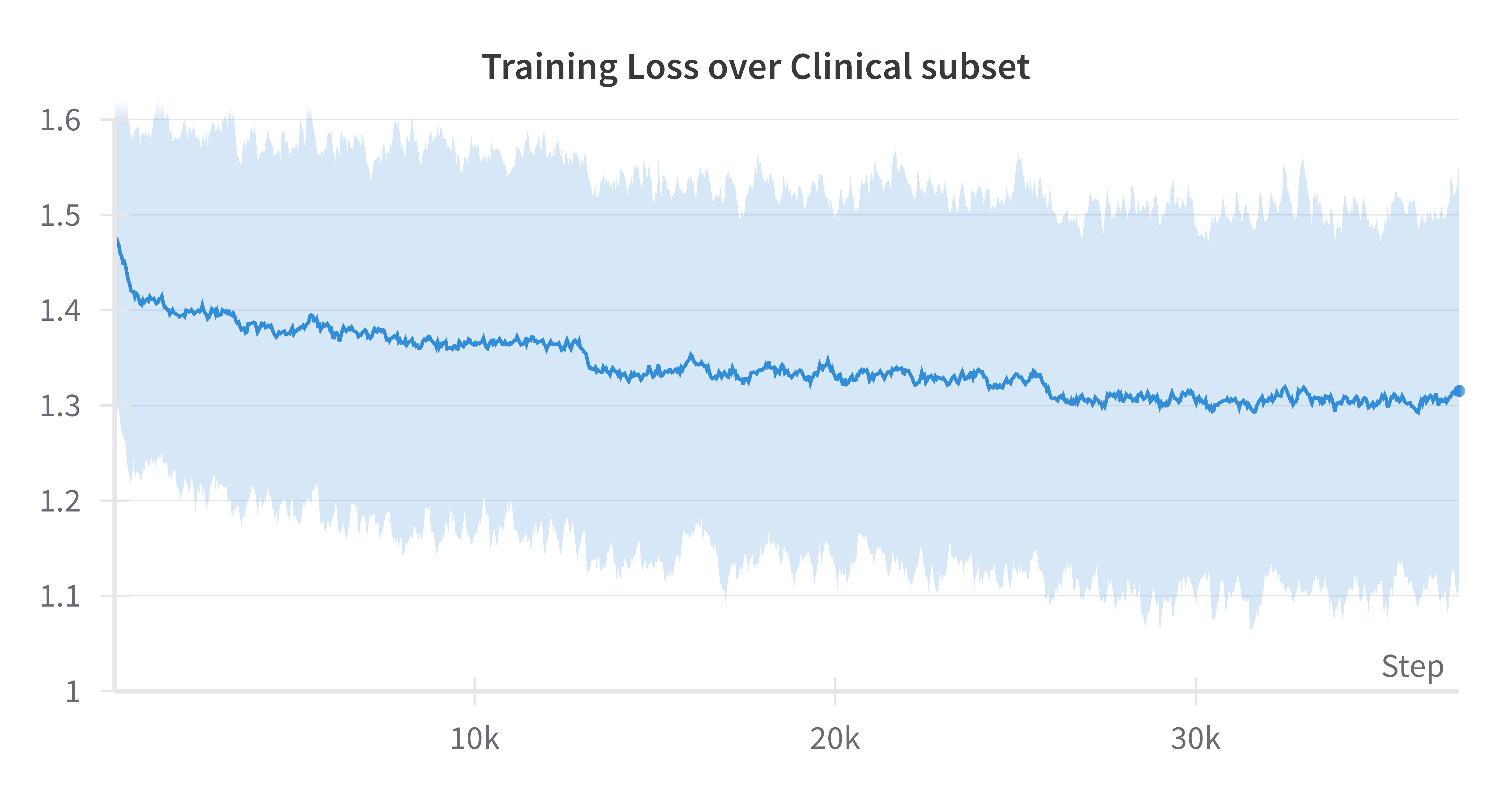}}
\caption{Training loss for Mixtral during continuous pretraining on the general (left) and clinical (right) subsets.}
\label{tab:loss_curves}
\end{figure*}

Following the work presented in \cite{gupta2023continual}, we implement a learning rate warm-up strategy, gradually increasing the learning rate over 1\% of the total training steps. Specifically, we employ a linear warm-up, starting from 1/10th of our maximum learning rate and gradually ramping up to the full value. This gradual increase helps stabilize the training process and prevents drastic updates to the model's weights early on. Second, we address the potential distribution shift by blending our specialized clinical data with general language data from SlimPajama \cite{cerebras2023slimpajama}. This curated blend results in a 65-billion-token dataset, comprising 50 billion tokens of specialized clinical data and 15 billion tokens of general language data. We then perform continuous pretraining on this dataset for a total of 4 epochs, processing 260 billion tokens and allowing the model to acquire domain-specific knowledge while retaining its proficiency in general language understanding. In Figure \ref{tab:loss_curves}, we illustrate the training loss curves over both the general and clinical data subsets. As depicted, our warm-up strategy and data mixture effectively mitigate instabilities, demonstrating smooth convergence and a steady decrease in loss throughout the training process. This approach ensures the model's overall capabilities remain robust and facilitates the acquisition of specialized clinical knowledge.

\subsection{Instruct Fine-Tuning}

Instruct fine-tuning is a technique that aims to align large language models (LLMs) with human intentions and preferences by training them on a dataset of instructions and their corresponding desired outputs. This approach enables LLMs to better understand and respond to user prompts, improving their ability to generate relevant and useful responses in a variety of tasks.

To facilitate effective learning from instructions, we adopt a structured format incorporating the keywords <|system|>, <|prompter|>, and <|assistant|>. This format explicitly delineates the roles of the system, the user providing the prompt, and the assistant generating the response. By clearly defining these relationships, we guide the model to better understand the intent behind instructions and generate appropriate, medically relevant outputs.

Each sample in our instruction-tuning dataset is composed of three elements: a system prompt, a user prompt, and the corresponding model response. To maximize the utilization of the model's available context length during training, we concatenate these samples across the entire dataset. The training process is auto-regressive and the loss is solely focused on the tokens comprising the responses. This targeted training strategy prioritizes the model's ability to generate accurate and relevant answers, rather than focusing on replicating prompts.

We train our models for 3 epochs using a cosine learning rate scheduler, which gradually decreases the learning rate over the course of training.

\subsection{NEFTune}

NEFTune, a novel instruction fine-tuning technique introduced in \cite{jain2023neftune}, offers an alternative approach to our traditional pipeline. This method involves injecting noise into the embedding layer during training, a process that has shown improvements in the quality of the model's output generation. Furthermore, the introduced noise during training could act as a regularization method to stabilize the learning process. The noise vector is created by independently sampling each entry from a uniform distribution within the interval $[-1, 1]$. This vector is then scaled by a factor determined by the tunable parameter $\alpha$, the sequence length $L$, and the embedding dimension $d$:

$$
X'_{\text{emb}} \leftarrow X_{\text{emb}} + \left( \sqrt{\frac{\alpha}{Ld}} \right) \epsilon
$$

During our experiments, we explored various values for $\alpha$ and discovered that the setting of $\alpha = 5$ yielded superior results. In our study, we explore NEFTune as a potential replacement for our standard instruct fine-tuning pipeline, investigating its impact on overall performance on clinical tasks.

\begin{table*}[t]
\centering
\label{tab:hyperparameters}
\resizebox{\textwidth}{!}{
\begin{tabular}{l|ccc|ccc}
\toprule
\multirow{2}{*}{Hyperparameter} & \multicolumn{3}{c}{Mistral 7B} & \multicolumn{3}{|c}{Mixtral 8x7B} \\
\cmidrule(lr){2-4} \cmidrule(lr){5-7}
 & Pretraining & Fine-tuning & NEFTune & Pretraining & Fine-tuning & NEFTune \\
\midrule
Learning Rate Scheduler & \multicolumn{3}{c|}{Linear Warmup - Cosine} & \multicolumn{3}{c}{Linear Warmup - Cosine}\\
Max Learning Rate & $7\times10^{-6}$ & $5\times10^{-6}$ & $5\times10^{-6}$ & $7\times10^{-6}$ & $1\times10^{-6}$ & $1\times10^{-6}$\\
Beta & \multicolumn{3}{c|}{(0.9, 0.95)} & \multicolumn{3}{c}{(0.9, 0.95)} \\
Alpha & - & - & 5 & - & - & 5 \\
Weight Decay & \multicolumn{3}{c|}{0.1} & \multicolumn{3}{c}{0.1} \\
Number of Steps & 79,259 & 6,089 & 6,089 & 37,344 & 2,589 & 2,589 \\
\bottomrule
\end{tabular}
}
\caption{Hyperparameters for Pretraining, Fine-tuning, and NEFTune on Mistral 7B and Mixtral 8x7B Models}
\end{table*}

\subsection{Prompt Engineering}
In-Context Learning refers to a model's ability to understand and generate relevant responses based on the context provided within a prompt. This capability allows the model to leverage previous examples or instructions given in the prompt to perform tasks more effectively without explicit training on new samples. Chain-of-Thought Reasoning \cite{wei2023chainofthought} is a technique where the model is guided to generate a step-by-step explanation of its thought process before arriving at an answer. This approach encourages the model to articulate its reasoning, leading to more transparent and accurate outcomes.  In our work, we harness these capabilities by implementing the `Medprompt' prompting strategy as introduced by \cite{nori2023can}. To thoroughly evaluate our models, we generate chain-of-thought explanations using four distinct prompt engineering methods:
\begin{itemize}
    \item \textbf{Chain-of-Thought (CoT):} Similar to \cite{kojima2023large}, we generate chain-of-thought on the evaluation dataset by appending ``Let's think step-by step'' to every sample. This method encourages the model to systematically break down its thought process, leading to more structured and transparent reasoning.
    \item \textbf{Few-shot Chain-of-Thought:} In this approach, we improve the model's performance by providing context through static examples. Before generating the chain-of-thought explanation, we prepend the samples with five predefined few-shot examples. These examples serve as a guide, helping the model to understand and apply a consistent reasoning pattern.
    \item \textbf{Dynamic few-shot Chain-of-Thought:} This advanced method combines dynamic retrieval and chain-of-thought generation. Initially, we create chain-of-thought reasoning for multiple-choice question-answering datasets and store these in a Milvus vector database \cite{2021milvus}. We then embed the training questions using gte-small embedding model \cite{li2023towards}. During evaluation, we retrieve the five most semantically similar training examples based on cosine similarity in the embedding space. These retrieved examples are used as few-shot examples, providing relevant context to the model for generating more accurate explanations.
    \item \textbf{Dynamic few-shot Chain-of-Thought ensemble (CoT-En):} Building on the dynamic few-shot approach, this method introduces variability and robustness. Here, we shuffle the few-shot examples and the multiple-choice options, generating the chain-of-thought reasoning five times with a temperature setting of 0.2. This ensemble technique aims to produce a diverse set of reasonings.
\end{itemize}

\subsection{Hardware infrastructure.}
Our experiments were conducted on a high-performance computing cluster, utilizing a maximum of 10 nodes, each equipped with 8 NVIDIA H100 GPUs, for the continuous pretraining phase. For the subsequent fine-tuning stages, we employed 4 nodes of the same configuration. To efficiently train our large-scale models, we leveraged PyTorch's Fully Sharded Data Parallel (FSDP) \cite{zhao2023pytorch} framework, which enables distributed training across multiple GPUs while minimizing memory footprint. Additionally, we employed bfloat16 precision throughout our training pipeline.

\section{Datasets}
In this section, we detail our approach to constructing both the pretraining and fine-tuning datasets. Our primary objective is to curate datasets that optimize model performance while maintaining the highest standards of quality and relevance to the clinical domain. 

\subsection{Pretraining Dataset}




Our pretraining corpus comprises a mix of biology and healthcare data from publicly available sources, including full-text research articles, abstracts, open textbooks, and Wikipedia articles. We excluded data containing personally identifiable information as well as data without a permissive license for commercial use.

Pretraining data for large language models (LLMs) typically requires several normalization and cleaning steps to make it suitable for training. However, since we have controlled the input sources and limited them to trusted sources, our pretraining pipeline primarily involves five major steps: 1) document parsing, 2) low-length filtering, 3) document-level deduplication, 4) exact deduplication, and 5) data chunking.

Document parsing involves either scraping webpages or extracting text from research articles. Once the text is extracted from all sources, we remove sources with insufficient information by applying a length threshold filter. As our data mix mainly consists of full-text research articles, there is a high likelihood of document-level duplication with different DOI IDs. To address this, we used the MinHash \cite{broder1997minhash} deduplication technique with a similarity threshold of 0.85: for each document, we compute a sketch and measure its approximate similarity with other documents, removing pairs with high overlap. We perform MinHash deduplication using 9,000 hashes per document, calculated over 5-grams and divided into 15 buckets of 400 hashes each.

Document-level deduplication removes similar documents across different data sources, but there could still be some text duplication within the documents. Therefore, we additionally employed an exact deduplication step \cite{lee2021deduplicating} to eliminate identical text segments from the dataset. As advised in the original literature, we ran the exact deduplication twice with length thresholds of 400 and 100 bytes, since duplicates may persist even after the first pass. Finally, the entire dataset is tokenized, concatenated, and split into chunks with a predefined context length for continuous pretraining. 


\subsection{Finetuning Dataset}

Our instruction-tuning dataset is a curated blend of open-source medical question-answering data, sourced primarily from medical forums like Stack Exchange, rich in expert discussions and patient inquiries. We also integrate relevant medical segments extracted from general domain datasets, ensuring a diverse representation of medical subfields and contexts. This comprehensive dataset provides a solid foundation for training our model to accurately understand and generate medically relevant content.

To improve the chain-of-thought capabilities of the fine-tuned model, we generate chain-of-thought explanations for datasets that benefit from reasoning chains. After generating these reasoning chains, we discard those that do not correspond with the correct answers and use these samples as zero-shot examples. We employ the Mixtral-Instruct model for both generating and verifying the reasoning chains. For more details on the composition of the finetuning dataset, please refer to \autoref{tab:traindata}.

\section{Evaluations}

\begin{table*}[h!]
    \centering
    \begin{tabular}{c|c|ccccc}
         & \# of parameters & MedQA & USMLE & MMLU  & MedMCQA  \\ \hline 
         BioMistral \cite{labrak2024biomistral} & 7B & 45.09 & 46.67 & 63.63 & 44.58 \\
         Clinical Camel \cite{toma2023clinical} & 70B & 53.42 & 54.35  & 69.75 & 47.01 \\
         MediTron \cite{chen2023meditron} & 70B & 51.14 & 57.31 & 68.26 & 42.36 \\
         Med42 \cite{christophe2024med42} & 70B & 61.52 & 72.01 & 76.71 & 60.93 \\ \hline\hline
         Mistral 7b Instruct & 7B & 42.89  & 48.18 & 62.75 & 43.32 \\
         Mistral 7b $F$ (ours) & 7B & 54.28 & 62.63 & 68.30 & 58.11 \\
         Mistral 7b $N$ (ours) & 7B & 60.72 & 61.97 & 70.35 & 58.57 \\
         Mistral 7b $P+F$ (ours) & 7B & 58.36  & 63.84 & 72.28 & \textbf{60.84} \\
         Mistral 7b $P+N$ (ours) & 7B & \textbf{62.69} & \textbf{63.98} & \textbf{73.45} & 59.79 \\ \hline\hline
         Mixtral 8x7b Instruct & 46.7B & 52.55 & 65.99 & 75.78 & 53.74 \\
         Mixtral 8x7b $F$ (ours) & 46.7B & 62.60 & 72.68 & 79.10 & 62.85 \\
         Mixtral 8x7b $N$ (ours) & 46.7B & 66.93 & 70.05 & 79.57 & 64.64 \\
         Mixtral 8x7b $P+F$ (ours) & 46.7B & 67.09 & \textbf{73.57} & \textbf{79.92} & 65.29 \\
         Mixtral 8x7b $P+N$ (ours) & 46.7B & \textbf{68.34} & 72.82 & 79.84 & \textbf{65.34} \\ \hline
    \end{tabular}
    \caption{Accuracy over multiple clinical QA tasks. $F$ stands for Instruct-Finetuning, $P$ stands for Pretraining, and $N$ stands for NEFTune. We show that our models improve on all tasks as we gradually add more training techniques.}
    \label{tab:pretraining_finetuning_table}
\end{table*}

To rigorously assess the efficacy of our fine-tuning approaches, we focus on a comprehensive evaluation of the models' capabilities across a spectrum of clinical question-answering (QA) tasks. We employ a diverse suite of QA datasets, including MedQA \cite{jin2020disease}, USMLE sample exam and self-assessment \cite{nori2023capabilities, Han2023medalpaca}, MMLU (medical subset)\cite{hendryckstest2021}, and MedMCQA\cite{pmlr-v174-pal22a}, to ensure a thorough and representative assessment of model performance in various clinical scenarios.

Our evaluation methodology uses the EleutherAI Harness framework \cite{eval-harness}, which focuses on the likelihood of a model generating each proposed answer rather than directly evaluating the generated text itself. To enhance the granularity and relevance of our analysis, we introduce modifications to the Harness codebase. Instead of computing the likelihood of generating only the answer choice labels (a, b, c, or d), we extend the computation to encompass the likelihood of generating the complete answer text. This modification provides a more detailed understanding of the model's performance, as it takes into account the entire answer generation process, including the ability to articulate reasoning and justify the selected answer choice.

To evaluate the efficacy of MedPrompt prompting strategies, we integrate these prompts into the Harness framework. This involves generating reasoning chains based on the prompts and then using Harness to assess the likelihood of the model producing the final answer derived from these chains. This approach allows us to evaluate the impact of specific prompting techniques on the model's ability to reason through complex clinical scenarios.

Throughout our evaluation, we report accuracy as the primary metric across all tables, providing a clear and interpretable measure of the models' proficiency in clinical QA tasks.

\begin{figure*}[h!]
    \centering
    \includegraphics[width=\linewidth]{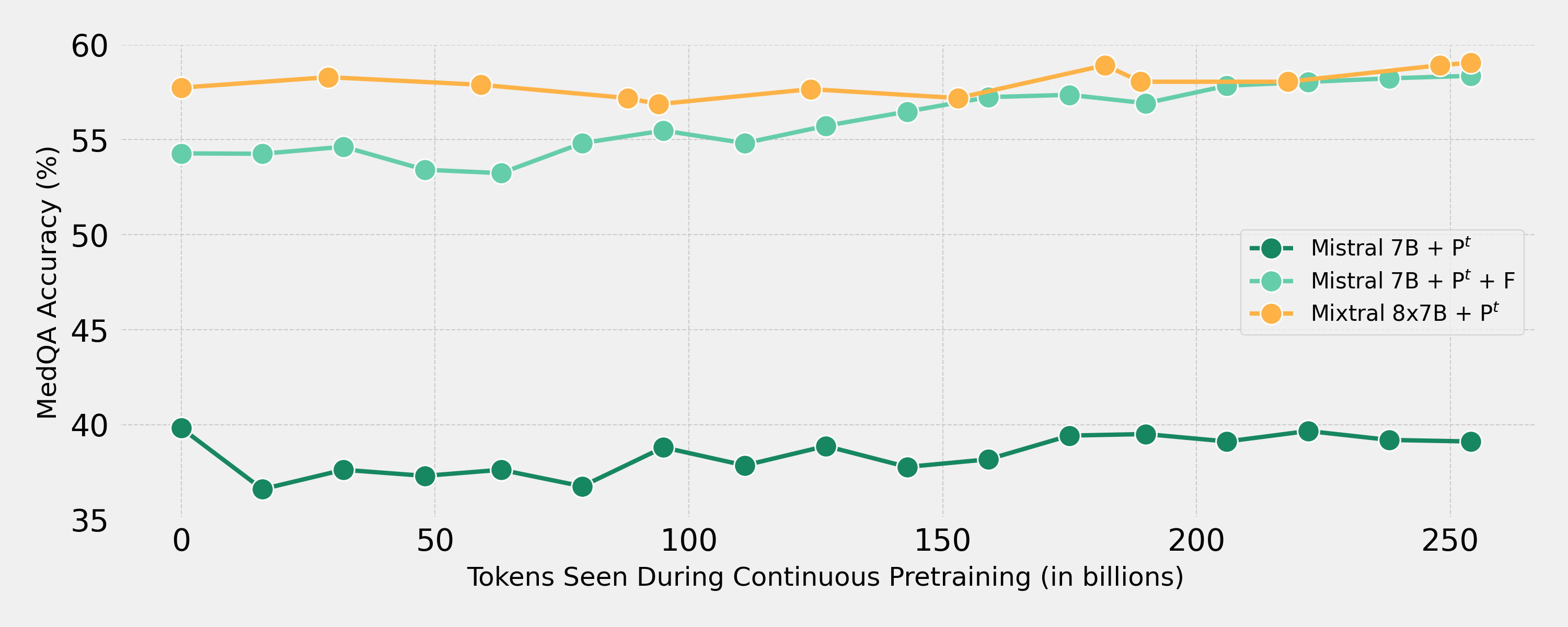}
    \caption{Evolution of MedQA accuracy for Mistral-7b and Mixtral 8x7b base models as well as our instructed versions of Mistral-7b during continuous pretraining. $Pˆt$: Continuous Pretrained with variable numbers of tokens $t$, $F$: Instruct Finetuned. We show that, while base model accuracy remains consistent, applying instruct-finetuning leads to notable improvements.}
    \label{fig:pretraining_finetuning}
\end{figure*}

\section{Results}

In this section, we present the results of our experiments, revealing key insights into the effectiveness of different training approaches for clinical language models.

\paragraph{Non-instructed models can't be evaluated on QA tasks.}  Throughout our continuous pretraining process, we saved multiple checkpoints and assessed their performance on our clinical QA benchmarks. Given the absence of instruction fine-tuning, we opted for a no-prompt evaluation format. As illustrated in Figure \ref{fig:pretraining_finetuning}, a slight performance decline is observed between the base model (with no pretraining) and the initial checkpoints. While subsequent checkpoints exhibit gradual improvement with increased exposure to clinical data, their performance consistently trails behind the original base model. This observation underscores the critical role of instruction fine-tuning in equipping LLMs with the necessary skills to effectively comprehend and respond to questions in the clinical domain.

\paragraph{Instruct Fine-tuning Specializes the Model for QA Data} The remarkable leap in performance observed in Table \ref{tab:pretraining_finetuning_table} after instruct fine-tuning highlights the efficacy of this approach in aligning LLMs with the specific demands of clinical question-answering tasks. While this outcome is not surprising or novel, it reaffirms the established effectiveness of fine-tuning methodologies in adapting models to specific domains. By exposing the models to a curated dataset of instructions and corresponding answers, we effectively specialize both Mistral and Mixtral to formulate answers in the clinical domain. This targeted training approach enhances the models' ability to understand the intent behind questions, use the knowledge acquired during pretraining, and generate accurate, relevant, and informative responses. The substantial gains observed across all benchmarks show the critical role of instruct fine-tuning in bridging the gap between general language understanding and specialized clinical expertise, ultimately empowering LLMs to excel in medical question answering.

\begin{figure*}[h!]
    \centering
    \includegraphics[width=\linewidth]{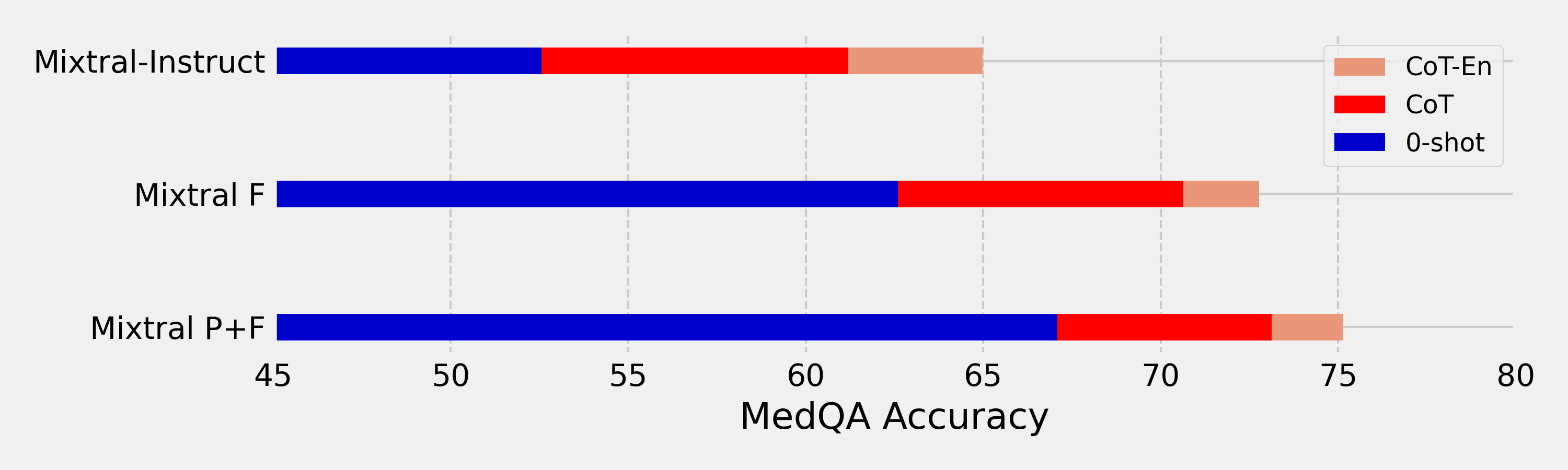}
    \caption{Evolution of MedQA accuracy using MedPrompt over different versions of Mixtral.}
    \label{fig:medprompt}
\end{figure*}

\paragraph{Continuous Pretraining Shows Consistent Performance Gains} To assess the impact of continuous pretraining on downstream performance, we conducted a comprehensive evaluation by instruct-fine-tuning various checkpoints saved during the pretraining process. Figure \ref{fig:pretraining_finetuning} illustrates the performance trajectory of these models as they are exposed to increasing amounts of pretraining data. Initially, the gains are relatively minor, particularly within the first 100 billion tokens. However, as the models continue to learn from the vast corpus of clinical text, we observe a gradual and steady improvement in their performance across a range of QA benchmarks.

This trend suggests that continuous pretraining serves as a valuable foundation, gradually enhancing the model's understanding of clinical concepts and terminology. As the models assimilate more domain-specific knowledge, they become better equipped to leverage the instruction data during fine-tuning, ultimately leading to superior performance on clinical QA tasks.

Table \ref{tab:pretraining_finetuning_table} provides a detailed breakdown of the performance gains achieved through continuous pretraining across various benchmarks. Notably, our continuously pretrained models consistently outperform state-of-the-art models, including the instruct-tuned versions of both Mistral and Mixtral. These results underscore the efficacy of continuous pretraining in equipping LLMs with the necessary domain knowledge to excel in clinical applications.

The magnitude of these gains, however, varies across model sizes. Mistral-7B demonstrates significant improvement, while the larger Mixtral 8x7B model exhibits more marginal, yet still consistent, benefits. This suggests that while continuous pretraining remains valuable for larger models, its impact may be less pronounced compared to smaller counterparts, potentially due to the diminishing returns of additional data for already extensive architectures. These findings demonstrate the importance of carefully weighing the computational costs and performance benefits of continuous pretraining, particularly for larger LLMs.

\paragraph{Adding noise helps finetuning.}

In our experiments, we observed an intriguing phenomenon with the NEFTune technique, originally proposed in \cite{jain2023neftune}. While the authors demonstrated that NEFTune applied to LLaMA-2 7B maintained Harness accuracy across several QA tasks, we show in Table \ref{tab:pretraining_finetuning_table}, that for Mistral-7B, it not only preserved but, in most cases, even improved the model's performance.  This performance increase was consistent across both the base model and the continuously pretrained model. This result is particularly surprising as NEFTune was primarily designed to enhance generation quality, not necessarily benchmark accuracy. We hypothesize that the injection of noise during training might act as a form of regularization, preventing overfitting and leading to better generalization on downstream tasks. However, the exact mechanisms behind this improvement warrant further investigation. This result suggests that the benefits of NEFTune extend beyond its intended purpose, potentially influencing the model's ability to reason and select the most likely answer.

\paragraph{Prompt Engineering makes the difference.}

Figure \ref{fig:medprompt} showcases the potential of MedPrompt as a viable alternative to traditional fine-tuning and pretraining techniques. By incorporating Chain-of-Thought (CoT) prompting and KNN CoT ensembles, we achieve substantial performance gains for the Mixtral-Instruct model on various clinical QA tasks. The effectiveness of MedPrompt is consistently observed across different model configurations: fine-tuned models outperform their non-fine-tuned counterparts, and pretraining followed by fine-tuning further amplifies these improvements. Notably, by employing MedPrompt with CoT and KNN CoT ensembles, we elevate MedQA accuracy from 52.55\% with the baseline Mixtral-Instruct model to a value exceeding 75\%. These results not only show the potential of advanced prompting strategies like MedPrompt to significantly enhance LLM performance in clinical applications without requiring computationally expensive fine-tuning or pretraining procedures, but also highlight the crucial role of pretraining in establishing a strong foundation for further improvement.

\section{Conclusion and Discussions}

In this study, we have systematically investigated the impact of continuous pretraining on in-domain clinical data, in conjunction with instruct fine-tuning and advanced prompting strategies, on the performance of LLMs in clinical question-answering tasks. Our findings demonstrate that continuous pretraining, while yielding modest improvements compared to other techniques, remains a valuable tool for enhancing LLM performance in the clinical domain. While continuous pretraining can often be challenging due to instability issues, we have shown that by carefully balancing in-domain clinical data with general language data, we can effectively mitigate these challenges and achieve consistent performance gains.

Furthermore, we have demonstrated that the benefits of continuous pretraining extend beyond the initial training phase, as it lays a solid foundation for subsequent instruct fine-tuning and the application of complex prompting techniques like MedPrompt. The synergy between continuous pretraining and these additional methods results in state-of-the-art performance on a variety of clinical QA benchmarks, outperforming existing models like the instruct-tuned versions of Mistral and Mixtral.

Our research opens up several avenues for future exploration. Further ablation studies could examine the effects of different domain data sources, beyond the clinical realm, on LLM performance. Additionally, a more comprehensive analysis of the optimal data mix for continuous pretraining, including varying proportions of in-domain and general language data, could yield valuable insights for maximizing the benefits of this technique.

This study provides a comprehensive framework for optimizing clinical LLM performance. Our findings offer valuable insights for future research and development efforts aimed at leveraging LLMs to address challenges and opportunities presented by the healthcare domain.

\section{Limitations}

While our research offers valuable insights into optimizing clinical LLMs, it is not without limitations. Primarily, our study focused on a specific set of models (Mistral and Mixtral) and a limited number of clinical QA datasets. While we strive for diversity in our benchmark selection, the generalizability of our findings to other LLM architectures or clinical tasks remains an open question.

Additionally, the computational resources required for continuous pretraining, particularly for larger models, may pose a barrier for widespread adoption. Further investigation into more efficient pretraining methods could address this limitation.

Finally, while our evaluation framework provides a comprehensive assessment of model performance on QA tasks, it does not fully capture the nuances of real-world clinical applications, where factors like explainability, bias mitigation, and safety are paramount. Future research should explore these aspects in greater detail to ensure the responsible and effective deployment of LLMs in healthcare settings.

\bibliography{custom}
\bibliographystyle{acl_natbib}

\appendix
\newpage
\onecolumn
\section{Appendix: Supplementary Materials}

\subsection{Finetuning Dataset Mix}
\begin{table*}[!h]
\begin{tabular}{lccll}
\toprule
\textbf{Dataset}                   & \multicolumn{1}{l}{\textbf{\# Samples}} & \multicolumn{1}{l}{\textbf{Mixture ratio (\%)}} &  &  \\
\midrule
Medical domain &  &  &  \\ 
~~~MedMCQA \cite{pmlr-v174-pal22a} & 180,462 & 23.49 \\ 
~~~Medical Flashcards \cite{Han2023medalpaca} & 30,106 & 3.92 \\ 
~~~StackExchange \cite{h4stackexchange} & 64,246 & 8.36 \\ 
~~~MedQA (USMLE) \cite{jin2020disease} & 11,290 & 1.47 \\ 
~~~CORD-19 \cite{wang-etal-2020-cord} & 17,721 & 2.31 \\ 
~~~PubMedQA \cite{jin2019pubmedqa} & 499 & 0.06 \\ 
~~~HeadQA \cite{vilares-gomez-rodriguez-2019-head} & 2,657 & 0.35 \\ 
~~~MediQA \cite{Han2023medalpaca} & 1,950 & 0.25 \\ 
~~~SciQ \cite{SciQ} & 11,679 & 1.52 \\ 
~~~PubMed Causal \cite{Han2023medalpaca} & 2,169 & 0.28 \\ 
~~~OpenGPT & 66,026 & 8.59 \\ 
~~~MedQUAD \cite{BenAbacha-BMC-2019} & 14,553 & 1.89 \\ 
~~~MMLU \cite{hendryckstest2021} & 244 & 0.03 \\ 
~~~Niv2* \cite{wang2022supernaturalinstructions} & 11,447 & 1.49 \\ 
~~~Pubhealth \cite{kotonya2020explainable} & 9,804 & 1.28 \\ 
Total & 424,853 & 55.29 \\ \midrule
General domain &  &  &  \\ 
~~~SlimOrca T0 \cite{OpenOrca, sanh2022multitask} & 109,235 & 14.22 \\ 
~~~SlimOrca Flan \cite{OpenOrca, longpre2023flan} & 109,169 & 14.21 \\ 
~~~SlimOrca CoT \cite{OpenOrca, wei2022finetuned} & 74,172 & 9.65 \\ 
~~~Ultrachat \cite{ding2023enhancing} & 50,953 & 6.63 \\ 
Total & 343,529 & 44.71 \\ 
\bottomrule
\multicolumn{5}{l}{\small \textsuperscript{\textdagger} The following categories were included: ``academia", ``bioinformatics'', ``biology", ``cogsci", ``fitness", ``health".} \\
\multicolumn{5}{l}{\small \textsuperscript{\textdaggerdbl} Only samples in English were used. } \\
\multicolumn{5}{l}{\small \textsuperscript{\textdollar} The following subjects were included: ``anatomy", ``clinical knowledge", ``college medicine", ``medical genetics", } \\
\multicolumn{5}{l}{\small ``professional medicine", ``college biology", ``high-school biology", ``professional psychology", ``high-school psychology", } \\
\multicolumn{5}{l}{\small ``human sexuality", ``human aging", ``nutrition", and ``virology".} \\
\multicolumn{5}{l}{\small * Samples from 47 tasks (from over 1,000 tasks) related to science, healthcare and medicine were included. }
\end{tabular}
\caption{Summary of subsets of the data used for fine-tuning the models. Note that medical-domain data correspond to approximately 60\% of the entire dataset.}
\label{tab:traindata}
\end{table*}

\newpage
\onecolumn

\subsection{Prompt formats}
\begin{figure}[h]
\centering
\includegraphics[width=0.8\textwidth]{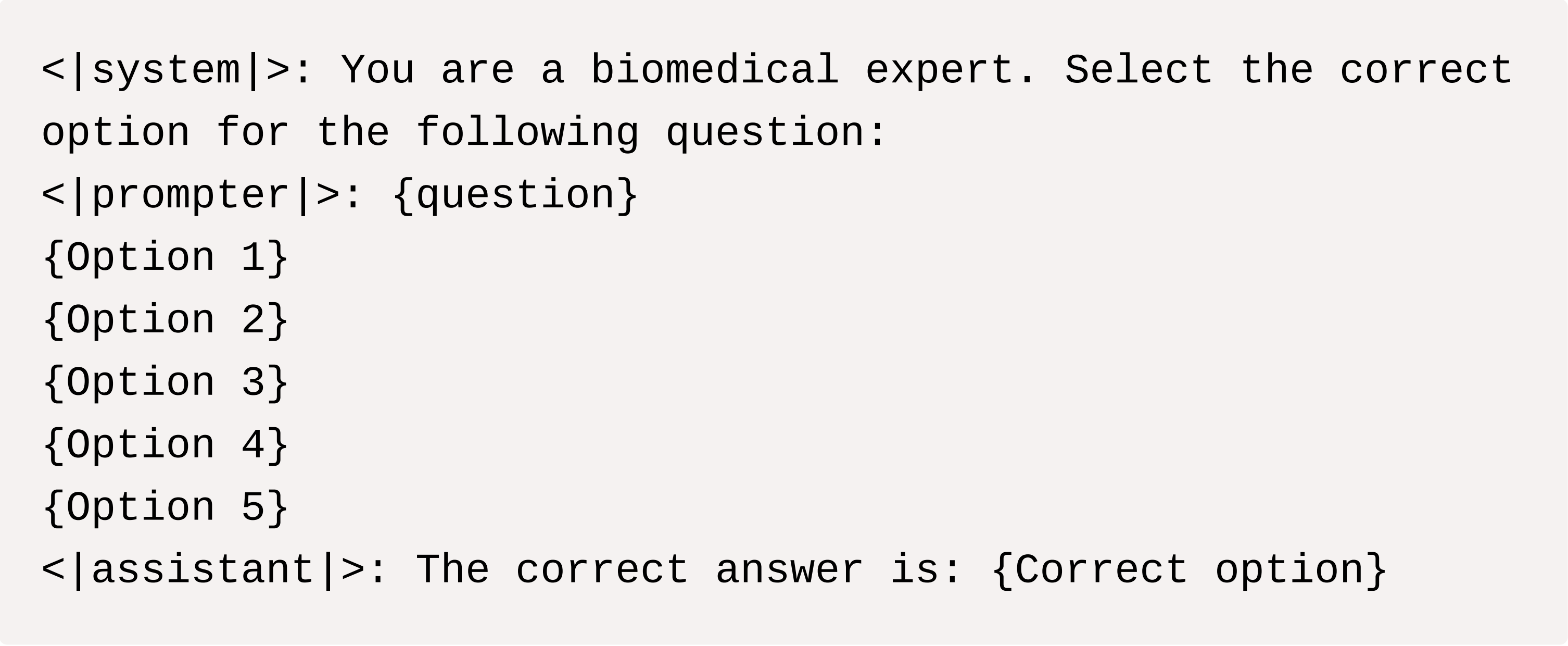}
\caption{Zero-shot prompt format on a sample from MedQA}
\label{fig:prompt_zero}
\end{figure}
\begin{figure*}[h]
\centering
\includegraphics[width=0.8\textwidth]{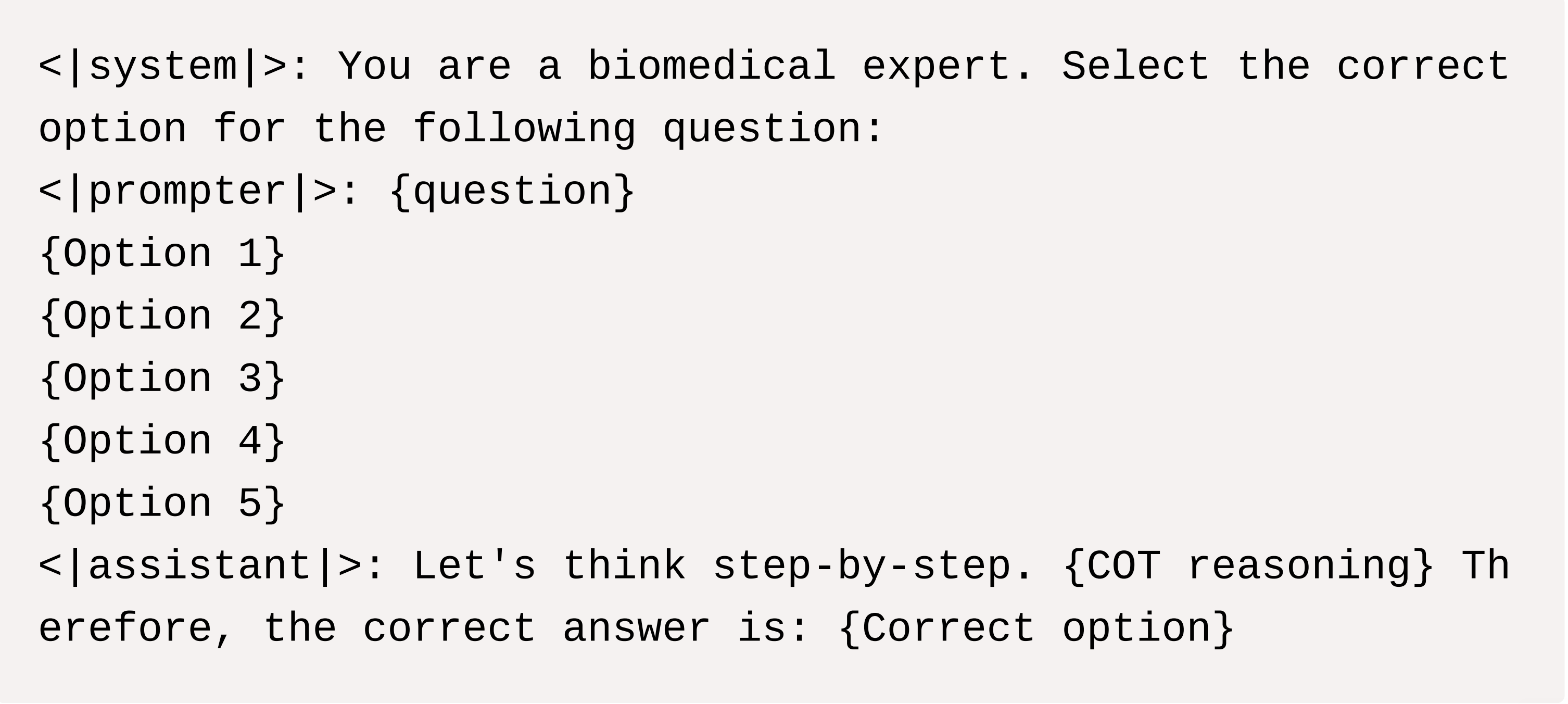}
\caption{Chain-of-Thought prompt format on a sample from MedQA}
\label{fig:prompt_cot}
\end{figure*}

\end{document}